\pdfoutput=1
\documentclass[10pt,twocolumn,letterpaper]{article}

\usepackage{iccv}              

\usepackage[accsupp]{axessibility}
\usepackage{amsmath}
\usepackage{afterpage}
\usepackage{amssymb}
\usepackage{booktabs}
\usepackage{pifont}
\usepackage{color}
\usepackage{bm}
\usepackage{multirow}
\usepackage{soul}
\usepackage{bbm}
\usepackage{subcaption}
\usepackage{xurl}
\usepackage{graphicx}
\usepackage{tabularx}  
\usepackage{listings}
\usepackage{xcolor}
\lstdefinestyle{promptstyle}{
    basicstyle=\ttfamily\footnotesize,
    backgroundcolor=\color{gray!10},
    frame=single,
    breaklines=true}

\usepackage[most]{tcolorbox}

\usepackage{longtable}
\usepackage{array}

\definecolor{iccvblue}{rgb}{0.21,0.49,0.74}
\definecolor{citecolor}{rgb}{0.21,0.49,0.74}
\usepackage[pagebackref=true,breaklinks=true,colorlinks,citecolor=citecolor,bookmarks=false]{hyperref}

\def\ours{\texttt{ByDeWay}}

\def\paperID{3} 
\def\confName{ICCV}
\def\confYear{2025}

\title{ByDeWay: Boost Your multimodal LLM with DEpth prompting in a training-free Way}

\author{
Rajarshi Roy\textsuperscript{1} \quad
Devleena Das\textsuperscript{1} \quad
Ankesh Banerjee\textsuperscript{1} \quad
Arjya Bhattacharjee\textsuperscript{1} \\
Kousik Dasgupta\textsuperscript{1} \quad
Subarna Tripathi\textsuperscript{2} \\ \\
\textsuperscript{1}Kalyani Government Engineering College, India 
\textsuperscript{2}Intel Labs, USA \\ \\
{\tt\small \{royrajarshi0123,devleena2003,banerjeeankesh\}@gmail.com} \\
{\tt\small arjyabhattacharjee5@gmail.com kousik.dasgupta@kgec.edu.in} \\
{\tt\small subarna.tripathi@intel.com}
}

\begin{document}

\maketitle


\begingroup
\renewcommand{\thefigure}{1}
\begin{figure*}[ht]
    \centering
    \includegraphics[width=\textwidth]{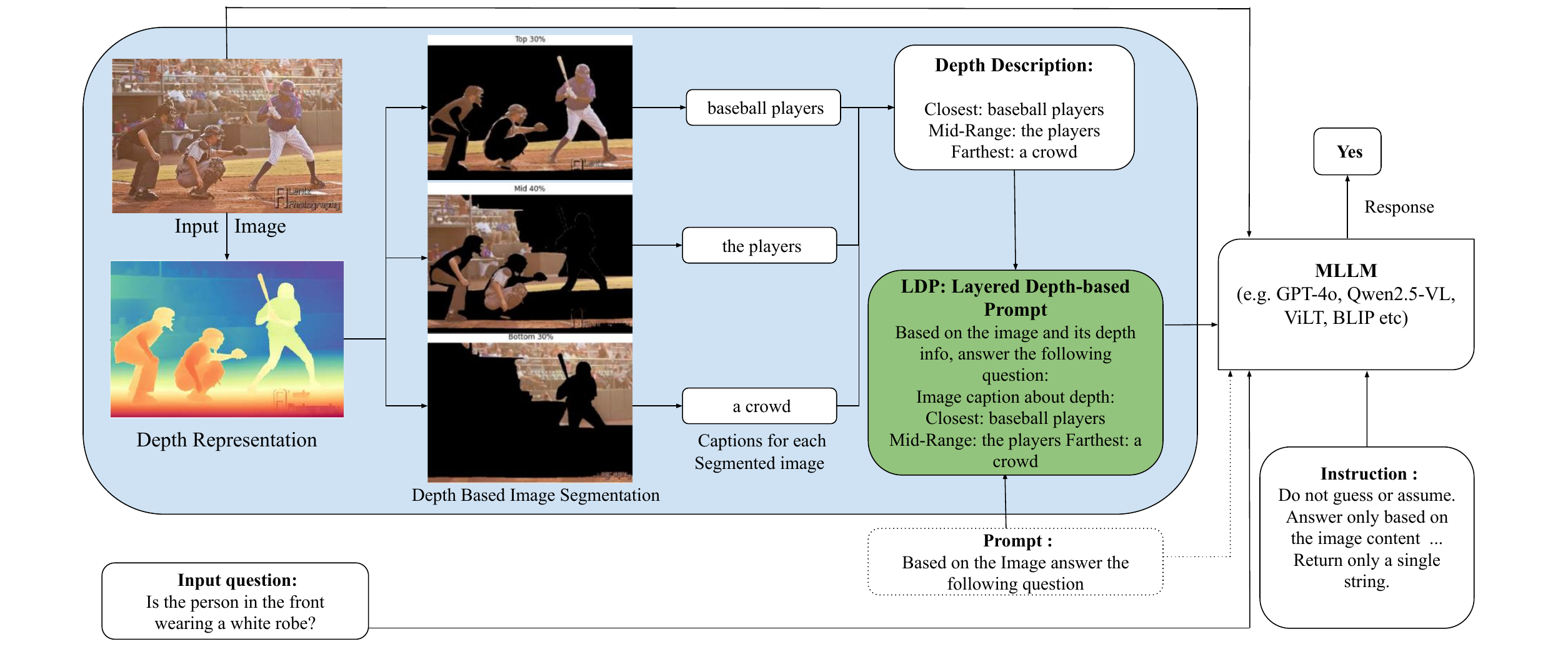}
    \caption{
    \textbf{\ours{} Workflow.} The light blue shaded area demonstrates the ByDeWay framework which utilizes Layered Depth-based Prompting (LDP) instead of standard prompting aiming to improve MLLM performances in a training-free way. The input image is processed to generate a depth map, which is then used to segregate visual information into layered regions (foreground, mid-ground, background). LDP augments standard prompting with the descriptions of such regions. MLLMs take input image, instruction, question and this augmented prompt as input and produces the response, the answer to the input question. 
    \textit{Please note that the standard prompt goes to the input, shown as a dotted line, to the MLLMs for our baseline comparisons.}
    }
    \label{fig:ldp_workflow2}
\end{figure*}
\endgroup

\begin{abstract}
We introduce a training-free framework, \textbf{ByDeWay}, a training-free method to boost the performances of Multimodal Large Language Models. Specifically, \emph{ByDeWay} leverages a novel prompting strategy, Layered-Depth-Based Prompting (LDP), that enhances the spatial reasoning and grounding capabilities of Multimodal Large Language Models (MLLMs). Our key insight is to inject structured spatial context derived from monocular depth estimation into the input prompts—without modifying any model parameters. By segmenting scenes into closest, mid-range, and farthest depth layers and generating region-specific captions using a grounded vision-language model, we produce explicit depth-aware textual descriptions. These descriptions are concatenated with image-question prompts to guide the model toward spatially grounded and hallucination-resistant outputs. Our method is lightweight, modular, and compatible with any black-box MLLM. Evaluations on hallucination-sensitive (POPE) and reasoning-intensive (GQA) tasks show consistent improvements across multiple MLLMs, demonstrating the effectiveness of depth-aware prompting in a zero-training setup.
The source code is available at 
\href{https://github.com/Rajarshi12321/ByDeWay}{https://github.com/Rajarshi12321/ByDeWay} .

\end{abstract}
  
\section{Introduction}

Multimodal Large Language Models (MLLMs) have emerged as powerful general-purpose systems capable of handling diverse tasks that span both language and vision modalities. From visual question answering (VQA) to image captioning and scene understanding, these models exhibit strong performance in zero-shot and few-shot settings by leveraging large-scale pretrained representations. Powerful MLLMs are unleashing new application potentials for digital marketing~\cite{pixie_MLLM} and advertisements~\cite{wang2025banneragencyadvertisingbannerdesign}. 


However, persistent challenges remain, chiefly \textbf{hallucination}—the tendency of MLLMs to generate content that is not grounded in the visual input—and \textbf{reasoning‑sensitivity}, where small changes in prompts or visuals can dramatically affect output. These issues become especially problematic in domains requiring precise spatial reasoning and object‑level grounding, such as robotics, healthcare, and safety‑critical applications. A recent survey finds that visual hallucinations—like misidentifying colors, objects, or spatial relations—continue to impede reliable deployment of MLLMs in real‑world settings \cite{bai2024hallucination}.

To address the above challenges, we present \textbf{ByDeWay} (\textbf{B}oost \textbf{Y}our multimodal LLM with \textbf{DE}pth prompting in a
training-free \textbf{Way}), a training-free method to boost the performances of Multimodal Large Language Models.

Rather than fine-tuning the model or altering its architecture, we propose a lightweight and modular prompting strategy: \textbf{LDP} (Layered-Depth-based Prompting). 




\textbf{LDP} explicitly incorporates depth‑aware textual context into the prompt by describing discrete spatial layers of the scene—foreground, mid‑range, and background—based on captions generated from monocular depth maps. These layer‑specific descriptions act as structured, interpretable anchors, capturing pseudo‑3D scene organization without any model retraining.

Concretely, we use an off‑the‑shelf monocular depth estimator to segment input images into depth‑based regions. Each region is captioned using a grounded vision–language model, yielding region‑specific textual summaries. These summaries are concatenated with the original image–question pair to form a \textbf{depth‑enriched prompt}, providing the MLLM with a clearer, spatially organized scene representation—thereby reducing hallucinations and bolstering visual reasoning in a fully \textbf{training‑free} manner (see Figure \ref{fig:ldp_workflow2}).

\vspace{2mm}
\noindent\textbf{Our key contributions are:}
\begin{itemize}
    \item We introduce \textbf{ByDeWay} (\textbf{B}oost \textbf{Y}our multimodal LLM with \textbf{DE}pth prompting in a training-free \textbf{Way}), a framework for boosting multimodal large language models, using the below  prompting strategy. 
    \item We propose \textbf{Layered-Depth-based Prompting (LDP)}, a novel prompting strategy that enriches MLLMs with explicit 3D spatial context using depth-aware captions.
    \item Our method is \textbf{modular and training-free}, requiring no fine-tuning of the base MLLM or additional supervision, making it scalable and model-agnostic.
    \item We demonstrate that LDP consistently improves performance on both hallucination-sensitive (POPE) and reasoning-intensive (GQA) benchmarks across multiple MLLMs.
\end{itemize}

\section{Related Work}

\subsection{Multimodal Prompting Techniques}
Similar to Large language models (LLM), reasoning capabilities of large multimodal models are improved with queries designed for targeted applications.   
Prompting in MLLMs has largely been around designing the textual queries. 
Researchers have focused on 
methods like CoT prompting \cite{wei2023chainofthoughtpromptingelicitsreasoning}, visual-context enrichment, and instruction tuning \cite{zhou2024visualincontextlearninglarge, wang2024m2ptmultimodalprompttuning}. 
Models like BLIP-2 \cite{li2023blip2bootstrappinglanguageimagepretraining} and Flamingo \cite{alayrac2022flamingovisuallanguagemodel} 
use both image and text inputs to adapt their prompts. 
While BLIP-2 utilize \emph{soft visual prompting}, Flamingo performs interleaved prompt generation. 
However, multimodal prompting, where the prompting modality is different than the image and the text, is still largely unexplored in academia, partially because most of the recent open-sourced models are of limited capacity. 
A thread of studies explore having MLLMs predicting object locations as bounding boxes~\cite{visionLLM23, peng2023kosmos2} or masks~\cite{hanoona2023GLaMM}. 
In recent work, set-of-marks~\cite{Somllava-yan2024list,yang2023setofmark} as a visual prompting 
notably improved the visual grounding abilities in the strongest MLLM.  
We introduce the concept of utilizing depth-based segmentation to enhance the prompt with spatial information.



\subsection{Depth-Aware Caption Generation and Hybrid Multimodal Systems}

Traditional image captioning models such as “Show and Tell” \cite{vinyals2015tellneuralimagecaption} and bottom-up attention \cite{anderson2018bottomuptopdownattentionimage} primarily rely on 2D visual features, overlooking depth and spatial reasoning. Recently, several works have demonstrated that incorporating depth information can significantly enhance spatial understanding in captions. For example, DEVICE \cite{Xu_2025} enriches the TextCaps task by integrating depth maps and visual-concept tokens via a depth-enhanced feature updating module, leading to more accurate relations between scene objects. Similarly, Ahmed et al.\ \cite{ahmed2023enhancingimagecaptioningdepth} propose a Transformer-based RGB–D captioning framework that explores pixel-, feature-, and hybrid-level fusion strategies on the NYU‑v2 dataset, showing improved multi-sentence descriptions by leveraging depth cues.

In parallel, dense captioning approaches operating on 3D point clouds go further by explicitly segmenting objects and reasoning about their inter-object relations. Scan2Cap introduced context-aware scene-object graphs and relational attention for dense descriptions of RGB–D scans \cite{chen2020scan2capcontextawaredensecaptioning}, while SpaCap3D \cite{wang2022spatialityguidedtransformer3ddense} improved on this by introducing a token-to-token spatiality-guided Transformer encoder that learns fine-grained 3D spatial relations among objects. X‑Trans2Cap \cite{yuan2022xtrans2capcrossmodalknowledgetransfer} further extends these ideas via cross-modal knowledge transfer, distilling RGB-aware features into a point-cloud-only model for efficient inference.

Complementing these captioning systems, numerous segmentation architectures demonstrate that depth and RGB fusion enhance scene decomposition. For instance, HybridNet performs joint depth estimation and semantic segmentation using shared backbone branches \cite{Sanchez_Escobedo_2018}, and Fooladgar \& Kasaei \cite{fooladgar2019multimodalattentionbasedfusionmodel} adopt attentional fusion of RGB–D for improved segmentation performance. Moreover, modular multimodal pipelines like Descriptive Caption Enhancement (DCE) integrate specialized depth, spatial, and interaction modules to generate richer and more structured captions \cite{sun2025descriptivecaptionenhancementvisual}.

Our work draws on these strands by explicitly segmenting scenes according to depth and seamlessly integrating a depth-informed segmentation module into a hybrid captioning pipeline. In contrast to prior 2D-only or loosely coupled RGB–D models, we introduce a unified framework that segments the scene along depth discontinuities and delivers captions grounded in 3D spatial organization. This contributes a novel fusion mechanism that bridges the gap between depth-aware segmentation and captioning."

\section{Methodology}

\subsection*{LDP: A Training-Free Modular Pipeline}

We introduce a structured, modular pipeline designed to enhance the spatial understanding of Multimodal Large Language Models (MLLMs) without any additional training or finetuning. Our approach repurposes depth cues as prompt-level information, enabling MLLMs to reason over spatially grounded representations. The pipeline consists of the following stages:

\begin{enumerate}
    \item \textbf{Image Acquisition:} A standard RGB image of a real-world scene is provided as input. This image serves as the basis for both visual and spatial reasoning. \\

    \item \textbf{Monocular Depth Estimation:} We employ an off-the-shelf monocular depth estimation model (Depth Anything V2) to compute a dense depth map $D(x, y)$, assigning a relative depth value to each pixel. The model operates in a zero-shot setting, requiring no supervision, and yields an affine-invariant inverse depth representation suitable for coarse spatial segmentation. \\

    \item \textbf{Depth-Based Segmentation:} Using the estimated depth map, we partition the image into three spatial layers:
    \begin{itemize}
        \item \textit{Closest Layer}: Top 30\% of depth values (nearest objects).
        \item \textit{Mid-Range Layer}: Middle 40\% (intermediate objects).
        \item \textit{Farthest Layer}: Bottom 30\% (background or distant regions).
    \end{itemize} 
    Thresholds $T_1$ and $T_2$ are computed at the 30th and 70th percentiles to define binary masks: $M_{\text{close}}, M_{\text{mid}}, M_{\text{far}}$. These masks isolate each layer from the full image, producing depth-segmented subregions. \\

    \item \textbf{Region-Aware Captioning via KOSMOS-2:} Each segmented region is passed individually to KOSMOS-2~\cite{peng2023kosmos2}, a grounded multimodal large language model. The model generates a localized caption describing only the visible content within the masked area. This produces three spatially grounded textual descriptions, one per depth layer. \\

    \item \textbf{Prompt Construction:} The layer-wise captions are concatenated into a structured, depth-aware prompt. This enhanced prompt is combined with the original task input (e.g., a question) and passed to the downstream MLLM. The inclusion of explicit spatial context guides the model's reasoning and discourages hallucination by anchoring its responses to structured visual evidence.
\end{enumerate}

This pipeline operates entirely in a \textbf{training-free and model-agnostic} manner. It treats the MLLM as a black-box and augments only its input space, making it scalable across diverse architectures and use cases.

\begin{figure}[htbp]
    \centering
    \includegraphics[width=\linewidth]{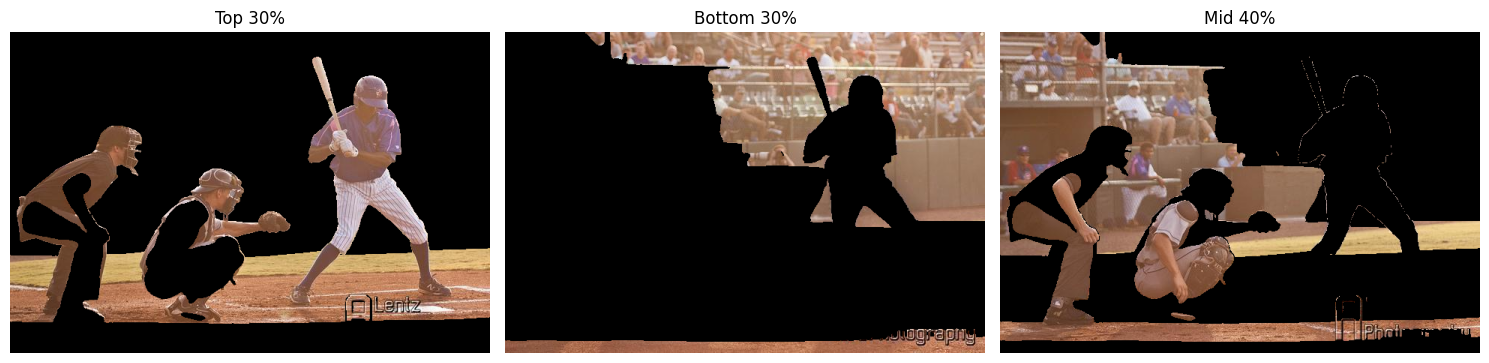}
    \caption{Depth-based segmented image showing the spatial separation of regions for an example input. The RGB image is divided into foreground (closest), mid-range, and background (farthest) layers using percentile-based depth thresholds.}
    \label{fig:depth-segmentation-example}
\end{figure}


\begin{figure}[t!]
    \centering

\begin{tcolorbox}[
    colback=gray!5!white, 
    colframe=gray!75!black, 
    title=LDP: Prompt with structured depth information, 
    boxrule=0.4pt,
    fonttitle=\bfseries,
    top=5pt,
    left=5pt,
    right=5pt,
    bottom=5pt
]
\footnotesize
\textbf{Based on the image and its depth information, answer the following question with either 'yes' or 'no':}

\vspace{5pt}
\textbf{Question:} Is the person in the front wearing a white robe?

\vspace{5pt}
\textbf{Image Caption about depth:}
\begin{itemize}
    \item \textcolor{blue}{Closest:} A baseball player
    \item \textcolor{blue}{Mid-range:} The players
    \item \textcolor{blue}{Farthest:} A crowd
\end{itemize}

\vspace{5pt}
\textbf{Instructions:}
\begin{itemize}
    \item Do not guess or assume. Answer only based on the image content and depth info.
    \item Return only a single 'yes' or 'no' string.
    \item Example: 'yes'
\end{itemize}
\end{tcolorbox}

\vspace{1mm}

\begin{tcolorbox}[
    colback=gray!5!white, 
    colframe=gray!75!black, 
    title=Baseline: Standard Prompting, 
    boxrule=0.4pt,
    fonttitle=\bfseries,
    top=5pt,
    left=5pt,
    right=5pt,
    bottom=5pt
]
\footnotesize
\textbf{Based on the image, answer the following question with either 'yes' or 'no':}

\vspace{5pt}
\textbf{Question:} Is the person in the front wearing a white robe?

\vspace{5pt}
\textbf{Instructions:}
\begin{itemize}
    \item Do not guess or assume. Answer only based on the image content.
    \item Return only a single 'yes' or 'no' string.
    \item Example: 'yes'
\end{itemize}
\end{tcolorbox}

\captionof{figure}{Prompting example from POPE. Bottom: baseline prompt without spatial guidance. Top: our method injects structured depth-aware context, leading to more grounded and accurate model predictions.}
\label{figure:LDP-example}
\end{figure}

To illustrate the effectiveness of depth-based prompting, we show a side-by-side comparison in a Visual Question Answering (VQA) scenario from the POPE benchmark. The prompt with depth-aware context allows the MLLM to better localize the queried object and answer accurately without guessing.


\section{Experiments}

\subsection{Evaluating \ours{} on hallucination mitigation and visual reasoning}
We conduct experiments to assess the effectiveness of \ours{} approach in enhancing the performance of MLLMs. Our evaluation spans two distinct and complementary tasks:
(1) mitigating hallucinations in binary visual question answering (VQA) using the POPE benchmark~\cite{pope_Li-hallucination-2023}, and 
(2) improving spatial and semantic reasoning using the GQA benchmark~\cite{hudson2018gqa}. These tasks are selected to demonstrate the generalizability and robustness of depth-aware prompting in both hallucination-sensitive and reasoning-intensive settings.

To ensure qualitative depth-caption generation and manual inspection, we conduct our experiments on a curated subset of \textbf{150 samples} from each dataset. This allows for controlled evaluation while maintaining diversity in visual content and question types. 
\subsection{Datasets and Task Overview}

\begin{itemize}
    \item \textbf{POPE:} This benchmark comprises image-question pairs specifically curated to evaluate hallucination tendencies in MLLMs. Each sample poses a binary ("yes"/"no") question where hallucinations often arise from references to nonexistent objects. For our study, we use a \textbf{subset of 150 examples} and augment the prompts with structured, depth-layered captions generated through the LDP pipeline to inject explicit spatial grounding.

    \item \textbf{GQA:} The GQA dataset tests models on fine-grained visual reasoning, particularly involving spatial relations, object attributes, and logical compositions. We evaluate LDP on a \textbf{subset of 150 examples} to investigate its impact on spatial understanding in a training-free setup.
\end{itemize}

\subsection{Prompt Variants}
We describe the details of our Layered Depth-based Prompting (LDP) below. 
To isolate the impact of depth information, we compare two prompting configurations:

\begin{enumerate}
    \item \textbf{LDP: Depth Enhanced Prompt:} The prompt includes the original image-based question, followed by our structured depth-aware caption. The model is explicitly instructed to rely only on the image and depth descriptions, without making external assumptions.
    
    \item \textbf{Baseline Prompt:} The model is presented with only the question and image, with no supplementary spatial context. This reflects the default behavior of most MLLMs.
\end{enumerate}

Both configurations standardize the output format and require the model to produce structured, deterministic responses, allowing reliable comparison.

\subsection{Metrics}

We report the following standard classification metrics:
\emph{Accuracy}, \emph{Precision}, \emph{Recall}, and \emph{F1 Score}. POPE results are computed using binary classification metrics, while GQA is evaluated using \emph{accuracy}.

\subsubsection*{Results and Analysis}

The performance across models and tasks is summarized in Tables~\ref{tab:POPE_result_table} and~\ref{tab:GQA_result_table}. Across all models and tasks, the addition of depth-aware context through LDP yields consistent improvements.

\begin{table}[t!]
    \centering
    \setlength{\tabcolsep}{5pt}
    \resizebox{\linewidth}{!}{
    \begin{tabular}{c|cccc}
    \toprule
       $Method$ & Accuracy & Precision & Recall & F1 Score   \\ \midrule
       gpt-4o	& 0.860	& 0.861	& 0.879	& 0.861 \\
       gpt-4o-LDP (\bf ours) & \bf 0.873	& \bf 0.875	& \bf 0.893	& \bf 0.874  \\ 
       \midrule
       Qwen2.5-VL	& 0.7267	& 1.0000	& 0.7267	& 0.8417 \\
       Qwen2.5-VL-LDP (\bf ours)	& \bf 0.9000	& \bf 1.0000	& \bf 0.9000	& \bf 0.9474 \\
       \midrule
       ViLT	& 0.8533	& 0.8674	& 0.8533 &	 0.8549\\
       ViLT -LDP (\bf ours)	& \bf 0.9267	& \bf 0.9319	& \bf 0.9267	& \bf  0.9273\\
       \midrule
       BLIP	& 0.8733	& 1.0000 & 0.8733 & 0.9300 \\
       BLIP-LDP (\bf ours)	& \bf 0.9533	& \bf 1.0000 & \bf 0.9533 & \bf 0.9800\\
       \bottomrule
    \end{tabular}
    }
    \caption{Results on the POPE dataset. LDP consistently improves MLLM performance on hallucination-sensitive binary VQA.}
    \label{tab:POPE_result_table}
\end{table}

\begin{table}[t!]
    \centering
    \setlength{\tabcolsep}{35pt}
    \resizebox{\linewidth}{!}{
    \begin{tabular}{c|c}
    \toprule
       $Method$ & Accuracy \\ \midrule
       Qwen2.5-VL	& 0.5007 \\
       Qwen2.5-VL-LDP (\bf ours)	& \bf 0.6592 \\
       \midrule
       ViLT	& 0.527 \\
       ViLT -LDP (\bf ours)	& \bf 0.627 \\
       \midrule
       BLIP	& 0.5552 \\
       BLIP-LDP (\bf ours)	& \bf 0.6704 \\
       \bottomrule
    \end{tabular}
    }
    \caption{Results on the GQA dataset. LDP enhances spatial reasoning and object comprehension in a training-free manner.}
    \label{tab:GQA_result_table}
\end{table}


\begin{figure*}
\begin{subfigure}[h]{0.5\linewidth}
\includegraphics[width=\linewidth]{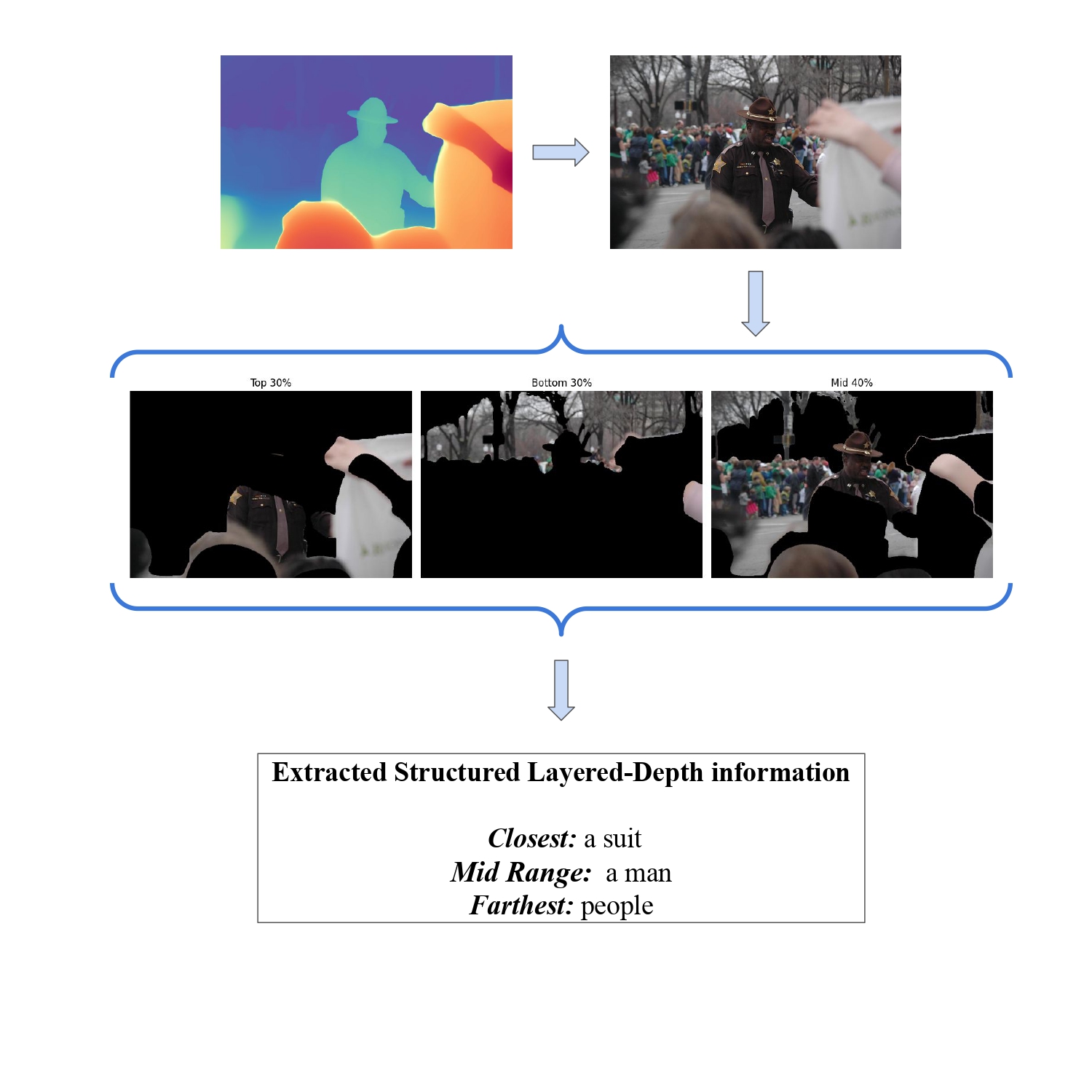}
\caption{LDP extraction: Example 1}
\end{subfigure}
\begin{subfigure}[h]{0.5\linewidth}
\includegraphics[width=\linewidth]{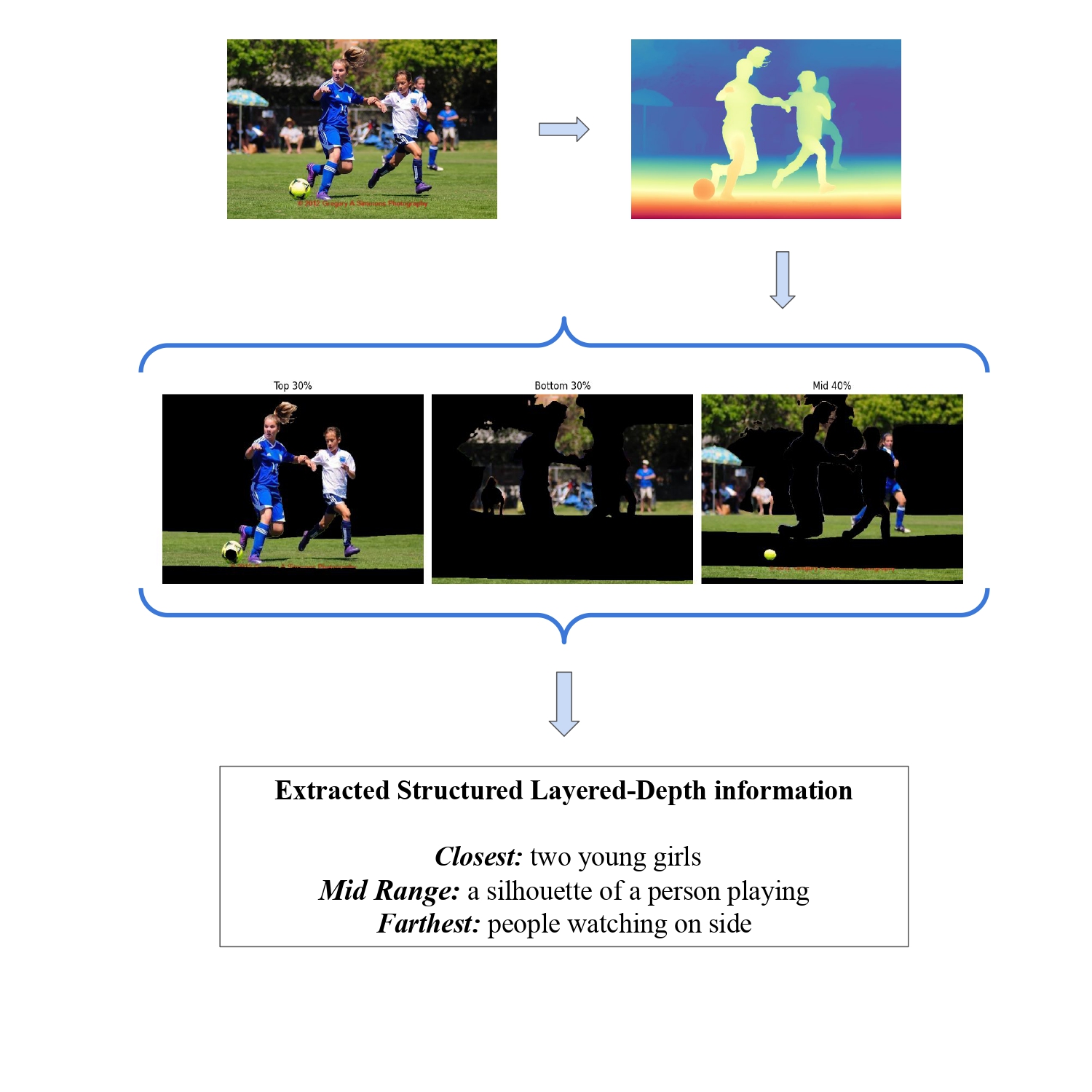}
\caption{LDP extraction: Example 2}
\end{subfigure}%

\begin{subfigure}[h]{0.5\linewidth}
\includegraphics[width=\linewidth]{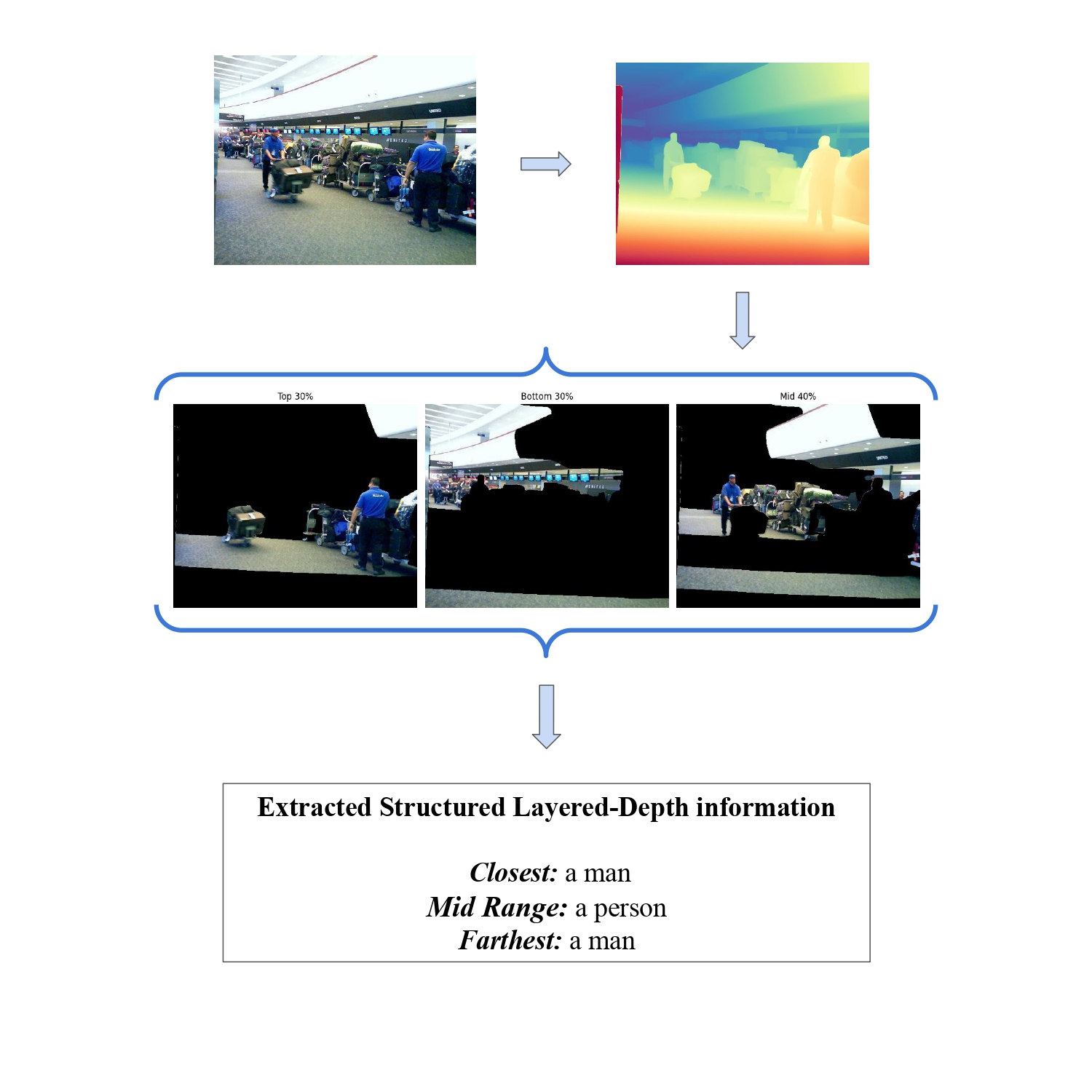}
\caption{LDP extraction: Example 3}
\end{subfigure}
\begin{subfigure}[h]{0.5\linewidth}
\includegraphics[width=\linewidth]{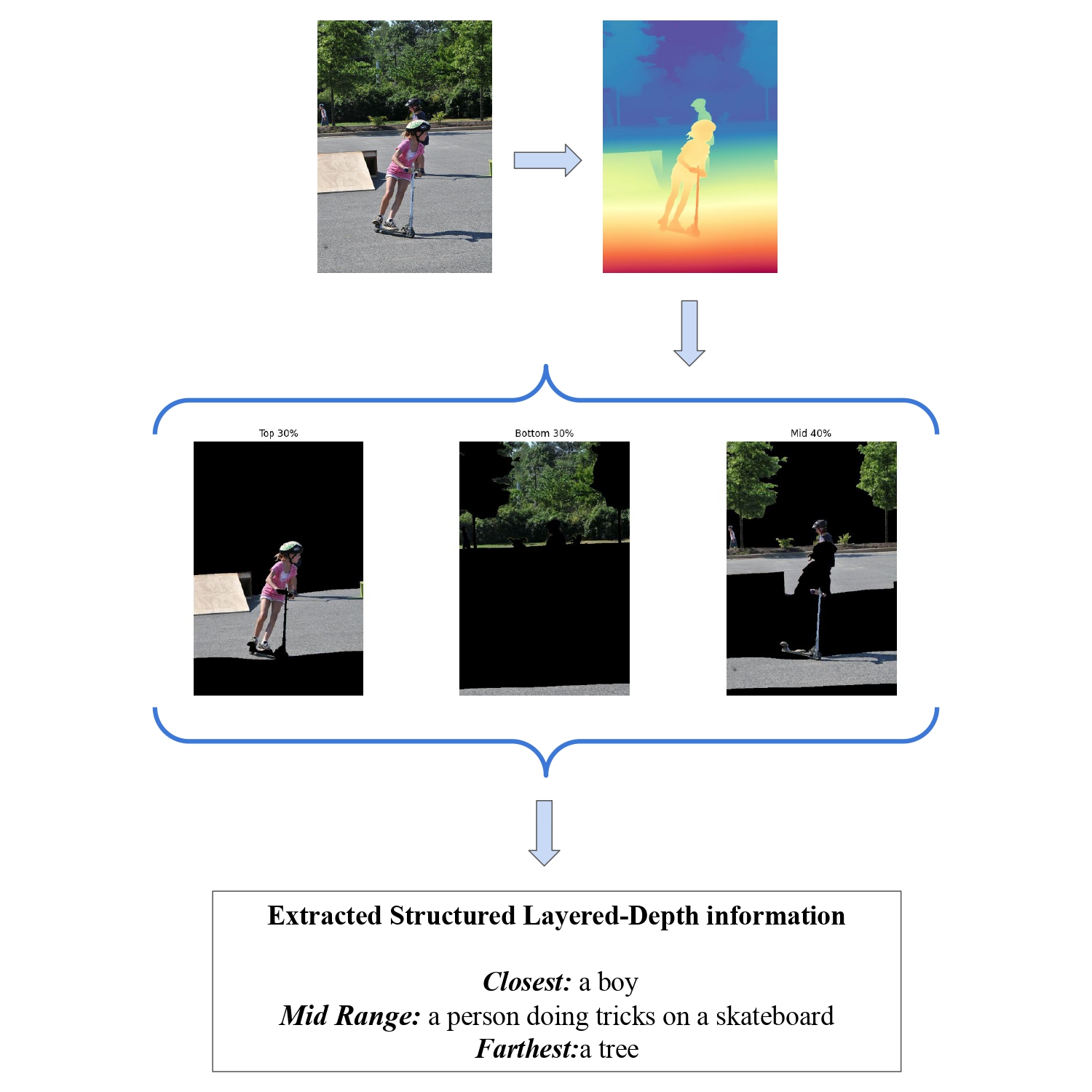}
\caption{LDP extraction: Example 4}
\end{subfigure}%

\caption{Qualitative examples of layered depth information extraction. From left to right and top to bottom in each group: Input image, extracted depth map, closest content, mid-range content and farthest content followed by their associated textual descriptions. }
\label{fig:LDP qualitative}
\end{figure*}

\subsubsection*{Insights and Implications}

In POPE, we observe consistent gains in Recall and F1 Score, indicating that LDP helps reduce false negatives and reinforces grounded answers. This is particularly valuable for hallucination-prone tasks, where spatial disambiguation is crucial. The improvements in BLIP and ViLT are especially notable, showing that even lighter-weight MLLMs benefit significantly from prompt-level spatial grounding.

In GQA, models augmented with LDP outperform their baselines by a substantial margin, affirming that even textual depth context alone can improve object reasoning and spatial understanding. Importantly, these improvements were achieved \textbf{without any model parameter updates}, confirming the value of structured visual semantics as a scalable, model-agnostic enhancement.

Our results confirm that LDP is a robust, plug-and-play method for improving MLLM performance in both hallucination-sensitive and reasoning-intensive tasks. By injecting depth-informed structure into prompts, we achieve substantial improvements in grounding and reliability—all in a training-free setup.
Figure~\ref{fig:LDP qualitative} shows examples of such layered depth information extraction qualitatively.

\section{Conclusions and Discussions}

We introduced \textbf{Layered-Depth-Based Prompting (LDP)}, a training-free technique for enhancing the spatial grounding capabilities of Multimodal Large Language Models (MLLMs). By injecting structured depth-aware textual descriptions—derived from monocular depth estimation and region-specific captioning—into the input prompt, LDP enables existing MLLMs to better interpret scene layout and reduce hallucination without modifying any model parameters.
Experimental results show that LDP boosts performances of several multimodal large language models on both hallucination-sensitive and reasoning-sensitive tasks in a training-free way, termed as \textbf{ByDeWay}. 
Importantly, LDP is model-agnostic and modular—it can be used with any black-box MLLM that accepts image-instruction inputs. This makes it a practical and scalable tool for real-world deployment, especially in scenarios where retraining large models is infeasible.

While LDP operates without additional training, it increases prompt length, which may affect inference latency in some settings. Future work could explore dedicated depth-token representations or multimodal adapters that compress layered-depth information more efficiently and effectively. 

To summarize, \textbf{ByDeWay}, our proposed framework, highlights that \emph{prompt-level spatial grounding}—powered by depth and layered captioning—is a powerful, training-free direction for improving the fidelity and robustness of modern multimodal systems.

{
    \small
    \bibliographystyle{ieeenat_fullname}
    \bibliography{egbib}
}

\end{document}